\documentclass{article}

\usepackage{amsthm,amssymb,amsfonts,amsmath}
\usepackage{graphicx}
\usepackage{caption}
\usepackage{subcaption}
\usepackage[ruled,linesnumbered]{algorithm2e}
\usepackage{authblk}

\usepackage[textsize=footnotesize,color=green!40]{todonotes}

\newtheorem{theorem}{Theorem}
\newtheorem{lemma}[theorem]{Lemma}
\newtheorem{corollary}[theorem]{Corollary}
\DeclareMathOperator*{\E}{\mathbb{E}}

\DeclareMathOperator{\diag}{\rm diag}

\newcommand{\R}{\mathbb{R}}
\renewcommand{\P}{\mathbb{P}}
\newcommand{\one}{\mathbb{I}}
\newcommand{\mat}[1]{\mathbf{#1}}

\newcommand{\rank}{\operatorname{rank}}
\newcommand{\trace}{\operatorname{Tr}}
\newcommand{\norm}[1]{\|#1\|}
\newcommand{\cX}{\mathcal{X}}
\newcommand{\cY}{\mathcal{Y}}
\newcommand{\cZ}{\mathcal{Z}}
\newcommand{\cE}{\mathcal{E}}

\newcommand{\hknormp}[1]{\norm{#1}_{\mathrm{H},p}}

\newcommand{\normop}[1]{\norm{#1}_{\mathrm{op}}}
\newcommand{\normfro}[1]{\norm{#1}_{\mathrm{F}}}
\newcommand{\normsch}[1]{\norm{#1}_{\mathrm{S},p}}
\newcommand{\normschq}[1]{\norm{#1}_{\mathrm{S},q}}
\newcommand{\normschtwo}[1]{\norm{#1}_{\mathrm{S},2}}
\newcommand{\normschinf}[1]{\norm{#1}_{\mathrm{S},\infty}}
\newcommand{\lone}{\ell_1}

\renewcommand{\H}{\mat{H}}
\newcommand{\sval}{\mathfrak{s}}
\newcommand{\vf}{\mat{f}}
\newcommand{\ve}{\mat{e}}
\renewcommand{\v}{\mat{v}}
\newcommand{\mR}{\mat{R}}
\newcommand{\mM}{\mat{M}}
\newcommand{\mV}{\mat{V}}

\newcommand{\fR}{\mathfrak{R}}
\newcommand{\hfR}{\widehat{\mathfrak{R}}}
\newcommand{\dUnif}{\mathbf{Unif}}
\newcommand{\cC}{\mathcal{C}}
\newcommand{\fL}{\mathfrak{L}}
\newcommand{\hfL}{\widehat{\mathfrak{L}}}

\newcommand{\sstar}{\Sigma^\star}
\newcommand{\cN}{\mathcal{N}}
\newcommand{\cF}{\mathcal{F}}
\newcommand{\cH}{\mathcal{H}}
\newcommand{\cHpr}{\mathcal{H}_{p,r}}
\newcommand{\cHtwor}{\mathcal{H}_{2,r}}
\newcommand{\cHoner}{\mathcal{H}_{1,r}}
\newcommand{\cR}{\mathcal{R}}
\newcommand{\cL}{\mathcal{L}}
\newcommand{\cRpr}{\mathcal{R}_{p,r}}

\newcommand{\cRtwor}{\mathcal{R}_{2,r}}
\newcommand{\cRoner}{\mathcal{R}_{1,r}}
\newcommand{\cA}{\mathcal{A}}
\newcommand{\cAnr}{\mathcal{A}_{n,p,r}}
\newcommand{\A}{\mat{A}}

\newcommand{\azero}{\boldsymbol{\alpha}}

\newcommand{\ainf}{\boldsymbol{\beta}}

\newcommand{\wa}{\langle \azero, \ainf, \{\A_a\} \rangle}
\newcommand{\wap}{\langle \azero', \ainf', \{\A_a'\} \rangle}

\newcommand{\wasigma}{\langle \azero, \ainf, \{\A_a\}_{a \in \Sigma} \rangle}

\newcommand{\ignore}[1]{}

\title{Generalization Bounds for Weighted Automata}

\author[1]{B.~Balle\thanks{Corresponding author: \texttt{b.deballepigem@lancaster.ac.uk}}}
\author[2,3]{M.~Mohri}
\affil[1]{Department of Mathematics and Statistics, Lancaster University}
\affil[2]{Courant Institute of Mathematical Sciences, New York University}
\affil[3]{Google Research}

\date{October 17, 2016}

\begin{document}

\maketitle

\begin{abstract}
  This paper studies the problem of learning weighted automata from a
  finite labeled training sample.  We consider several general
  families of weighted automata defined in terms of three different
  measures: the norm of an automaton's weights, the norm of the
  function computed by an automaton, or the norm of the
  corresponding Hankel matrix.  We present new data-dependent
  generalization guarantees for learning weighted automata expressed
  in terms of the Rademacher complexity of these families. We further
  present upper bounds on these Rademacher complexities, which reveal
  key new data-dependent terms related to the complexity of learning
  weighted automata.
%We also present an algorithmic framework for deriving regularized empirical risk minimization algorithms for weighted automata that can be analyzed using our techniques.
\end{abstract}

\section{Introduction}

Weighted finite automata (WFAs) provide a general and highly
expressive framework for representing functions mapping strings to
real numbers. The mathematical theory behind WFAs, that of rational power series, has been extensively studied in the past
\cite{Eilenberg1974,SalomaaSoittola1978,KuichSalomaa1986,BerstelReutenauer1988}
and has been more recently the topic of a dedicated handbook
\cite{DrosteKuich2009}. WFAs are widely used in modern applications,
perhaps most prominently in image processing and speech recognition
where the terminology of \emph{weighted automata} seems to have been
first introduced and made popular
\cite{CulikIIKari1993,MohriPereiraRiley1996,PereiraRiley1997,Mohri1997,MohriPereiraRiley2008},
in several other speech processing applications such as speech
synthesis \cite{Sproat1995,AllauzenMohriRiley2004}, in phonological
and morphological rule compilation
\cite{KaplanKay1994,Karttunen1995,MohriSproat1996}, in parsing
\cite{MohriPereira1998}, machine translation \cite{DeGispert2010}, bioinformatics
\cite{DurbinEddyKroghMitchison1998,AllauzenMohriTalwalkar2008},
sequence modeling and prediction \cite{CortesHaffnerMohri2004}, formal
verification and model checking \cite{baier2009model,AminofKupfermanLampert2011}, in optical character
recognition \cite{Breuel2008}, and in many other areas.

The recent developments in spectral learning \cite{hsu09,denis} have
triggered a renewed interest in the use of WFAs in machine learning,
with several recent successes in natural language processing
\cite{mlj13spectral,icml2014balle} and reinforcement learning
\cite{Boots:2009,Hamilton:2013}. The interest in spectral learning
algorithms for WFAs is driven by the many appealing theoretical
properties of such algorithms, which include their polynomial-time
complexity, the absence of local minima, statistical consistency, and
finite sample bounds \emph{\`a la} PAC \cite{hsu09}.  However, the
typical statistical guarantees given for the hypotheses used in
spectral learning only hold in the realizable case. That is, these
analyses assume that the labeled data received by the algorithm is
sampled from some unknown WFA.  While this assumption is a reasonable
starting point for theoretical analyses, the results obtained in this
setting fail to explain the good performance of spectral algorithms in
many practical applications where the data is typically not generated
by a WFA.  See \cite{cai2015} for a recent survey of algorithms for learning WFAs with a discussion of the different assumptions and learning models.

There exists of course a vast literature in statistical learning
theory providing tools to analyze generalization guarantees for
different hypothesis classes in classification, regression, and other
learning tasks.  These guarantees typically hold in an agnostic
setting where the data is drawn i.i.d.\ from an arbitrary
distribution.  For spectral learning of WFAs, an algorithm-dependent
agnostic generalization bound was proven in \cite{balle2012spectral}
using a stability argument. This seems to have been the first
analysis to provide statistical guarantees for learning WFAs in an
agnostic setting. However, while \cite{balle2012spectral} proposed a
broad family of algorithms for learning WFAs parametrized by several
choices of loss functions and regularizations, their bounds hold only
for one particular algorithm within this family.

In this paper, we start the systematic development of
algorithm-independent generalization bounds for learning with WFAs,
which apply to all the algorithms proposed in
\cite{balle2012spectral}, as well as to others using WFAs as their
hypothesis class. Our approach consists of providing upper bounds on
the Rademacher complexity of general classes of WFAs. The use of
Rademacher complexity to derive generalization bounds is standard
\cite{koltchinskii2000rademacher} (see also
\cite{bartlett2001rademacher} and \cite{mohri2012foundations}).  It
has been successfully used to derive statistical guarantees for
classification, regression, kernel learning, ranking, and many other
machine learning tasks (e.g.\ see \cite{mohri2012foundations} and
references therein). A key benefit of Rademacher complexity analyses
is that the resulting generalization bounds are data-dependent.

Our main results consist of upper bounds on the Rademacher complexity
of three broad classes of WFAs. The main difference between these
classes is the quantities used for their definition: the norm of the
transition weight matrix or initial and final weight vectors of a WFA;
the norm of the function computed by a WFA; and, the norm of the
Hankel matrix associated to the function computed by a WFA.  The
formal definitions of these classes is given in
Section~\ref{sec:classes}.  Let us point out that our analysis of the
Rademacher complexity of the class of WFAs described in terms of
Hankel matrices directly yields theoretical guarantees for a variety
of spectral learning algorithms. We will return to this point when
discussing the application of our results. As an application of our Rademacher complexity bounds we provide a variety of  generalizations bounds for learning with WFAs using a bounded Lipschitz loss function; our bounds include both data-dependent and data-independent bounds.

\paragraph{Related Work.}
To the best of our knowledge, this paper is the first to provide
general tools for deriving learning guarantees for broad classes of
WFAs.  However, there exists some related work providing complexity
bounds for some sub-classes of WFAs in agnostic settings.  The
VC-dimension of deterministic finite automata (DFAs) with $n$ states
over an alphabet of size $k$ was shown by \cite{Ishigami1997123} to be
in $O(k n \log n)$. This can be used to show that the Rademacher complexity of this class of DFA is bounded by $O(\sqrt{n k \log n /m})$. For probabilistic finite automata (PFAs), it was
shown by \cite{abe1992computational} that, in an agnostic setting, a
sample of size $\widetilde{O}(k T^2 n^2 / \varepsilon^2)$ is sufficient to
learn a PFA with $n$ states and $k$ symbols whose log-loss error is at
most $\varepsilon$ away from the optimal one in the class when the error is measured on all strings of length $T$.  New learning bounds on the Rademacher complexity of DFAs and PFAs follow as
straightforward corollaries of the general results we present in this paper.

Another recent line of work, which aims to provide guarantees for
spectral learning of WFAs in the non-realizable setting, is the
so-called low-rank spectral learning approach
\cite{kulesza2014low}. This has led to interesting upper bounds on the
approximation error between minimal WFAs of different sizes
\cite{kulesza2015low}. See \cite{bpp15} for a polynomial-time
algorithm for computing these approximations.  This approach, however,
is more limited than ours for two reasons.  First, because it is
algorithm-dependent. And second, because it assumes that the data is
actually drawn from some (probabilistic) WFA, albeit one that is
larger than any of the WFAs in the hypothesis class considered by the
algorithm.

The rest of this paper is organized as
follows. Section~\ref{sec:prelim} introduces the notation and
technical concepts used throughout. Section~\ref{sec:classes}
describes the three classes of WFAs for which we provide Rademacher
complexity bounds. The bounds are formally stated and proven in
Sections~\ref{sec:Anr}, \ref{sec:Rpr}, and \ref{sec:Hpr}. In
Section~\ref{sec:distparams} we provide additional bounds required for
converting some sample-dependent bounds from Sections~\ref{sec:Rpr}
and~\ref{sec:Hpr} into sample-independent bounds. Finally, the
generalizations bounds obtained using the machinery developed in
previous sections are given in Section~\ref{sec:bounds}.

\section{Preliminaries}
\label{sec:prelim}

\subsection{Weighted Automata, Rational Functions, 
and Hankel Matrices}

Let $\Sigma$ be a finite alphabet of size $k$.
Let $\epsilon$ denote the empty string and $\sstar$ the set of all
finite strings over the alphabet $\Sigma$. The length of
$u \in \sstar$ is denoted by $|u|$. Given an integer $L \geq 0$, we
denote by $\Sigma^{\leq L}$ the set of all strings with length at most
$L$: $\Sigma^{\leq L} = \{ x \in \sstar \colon |x| \leq L \}$. Given two strings $u, v \in \sstar$ we write $u v$ for their concatenation.

\begin{figure}[t]
\centering
\begin{subfigure}[b]{0.4\textwidth}
\centering
\includegraphics[height=7em]{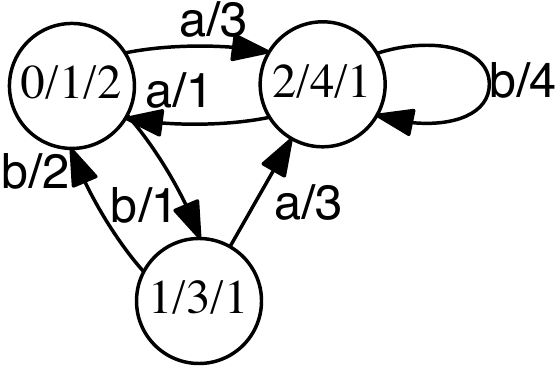}
\caption{\label{fig:trans}}
\end{subfigure}
\begin{subfigure}[b]{0.4\textwidth}
\centering
\begin{minipage}[t]{\textwidth}
\begin{tabular}{cc}
$\azero =
\left[
\begin{matrix}
1\\
3\\
4\\
\end{matrix}
\right]$
&
$\A_a =
\left[
\begin{matrix}
0 & 0 & 3\\
0 & 0 & 3\\
1 & 0 & 0\\
\end{matrix}
\right]$\\[.5cm]
$\ainf =
\left[
\begin{matrix}
2\\
1\\
1\\
\end{matrix}
\right]$
&
$\A_b =
\left[
\begin{matrix}
0 & 1 & 0\\
2 & 0 & 0\\
0 & 0 & 4\\
\end{matrix}
\right]$
\end{tabular}
\end{minipage}
\caption{\label{fig:weights}}
\end{subfigure}
\caption{(\protect\subref{fig:trans}) Example of WFA $A$. Within each circle, the first number indicates the state number, the second after the slash separator the initial weight and the third the final weight. In particular, $f_A(ab) = 1 \times 3 \times 4 \times 1 + 3 \times 3 \times 4 \times 1 + 4 \times 1 \times 1 \times 1$. (\protect\subref{fig:weights}) Corresponding initial vector $\azero$, final vector $\ainf$, and transition matrices $\A_a$ and $\A_b$.}
\label{fig:wfa}
%\caption{\protect\subref{fig:fig1} shows figure 1 and \protect\subref{fig:fig2} shows figure 2.}
\end{figure}

A WFA over the alphabet $\Sigma$ with $n \geq 1$ states is a tuple
$A = \wasigma$ where $\azero, \ainf \in \R^n$ are the initial and
final weights, and $\A_a \in \R^{n \times n}$ the transition matrix
whose entries give the weights of the transitions labeled with $a$.
Every WFA $A$ defines a function $f_A\colon \sstar \to \R$ 
defined for all $x = a_1 \cdots a_t \in \sstar$ by
\begin{equation}
\label{eqn:fAx}
f_A(x) = f_A(a_1 \cdots a_t) = \azero^\top \A_{a_1} \cdots \A_{a_t} \ainf =
\azero^\top \A_x \ainf \enspace,
\end{equation}
where $\A_x = \A_{a_1} \cdots \A_{a_t}$. This algebraic expression in fact corresponds to summing the weights of all possible paths in the automaton indexed by the symbols in $x$, where the weight of a single path $(q_0, q_1, \ldots, q_t) \in [n]^{t+1}$ is obtained by multiplying the initial weight of $q_0$, the weights of all transitions from $q_{s-1}$ to $q_{s}$ labeled by $x_{s}$, and the final weight if state $q_{t}$. That is:
\begin{equation*}
f_A(x) = \sum_{(q_0,\ldots,q_t) \in [n]^{t+1}} \azero(q_0) \left(\prod_{s = 1}^{t} \A_{x_s}(q_{s-1},q_s) \right) \ainf(q_t) \enspace.
\end{equation*}
See Figure~\ref{fig:wfa} for an example of WFA with $3$ states given in terms of its algebraic representation and the equivalent representation as a weighted transition diagram between states.

%\begin{center}
%\centering
%\begin{tabular}{c@{\hspace{1.25cm}}c}
%\includegraphics[scale=.66]{wfa} &
%\raisebox{1.cm}{
%\begin{minipage}[t]{0.3\textwidth}
%\begin{tabular}{cc}
%$\azero =
%\left[
%\begin{matrix}
%1\\
%3\\
%4\\
%\end{matrix}
%\right]$
%&
%$\A_a =
%\left[
%\begin{matrix}
%0 & 0 & 3\\
%0 & 0 & 3\\
%1 & 0 & 0\\
%\end{matrix}
%\right]$\\[.5cm]
%$\ainf =
%\left[
%\begin{matrix}
%2\\
%1\\
%1\\
%\end{matrix}
%\right]$
%&
%$\A_b =
%\left[
%\begin{matrix}
%0 & 1 & 0\\
%2 & 0 & 0\\
%0 & 0 & 4\\
%\end{matrix}
%\right]$
%\end{tabular}
%\end{minipage}}\\
%(a) & (b)
%\end{tabular}
%\caption{(a) Example of WFA $A$. Within each circle, the first
%  number indicates the state number, the second after the slash
%  separator the initial weight and the third the final weight. In
%  particular,
%  $f_A(ab) = 1 \times 3 \times 4 \times 1 + 3 \times 3 \times 4
%  \times 1 + 4 \times 1 \times 1 \times 1$.
%  (b) Corresponding initial vector $\azero$, final vector
%  $\ainf$, and transition matrices $\A_a$ and $\A_b$.}
%\label{fig:wfa}
%\end{center} 
%%\vskip -0.25in
%\end{figure}

An arbitrary function $f \colon \sstar \to \R$ is said to be \emph{rational} if there exists a WFA $A$ such that $f = f_A$. The \emph{rank} of $f$ is denoted by $\rank(f)$ and is defined as the minimal number of states of a WFA $A$
such that $f = f_A$. Note that minimal WFAs are not unique. In fact,
it is not hard to see that, for any minimal WFA $A = \wa$ with
$f = f_A$ and any invertible matrix $\mat{Q} \in \R^{n \times n}$,
$A^{\mat{Q}} = \langle \mat{Q}^\top \azero, \mat{Q}^{-1} \ainf,
\{\mat{Q}^{-1} \A_a \mat{Q}\} \rangle$
is also another minimal WFA computing $f$. We sometimes write $A(x)$
instead of $f_A(x)$ to emphasize the fact that we are considering a
specific parametrization of $f_A$. Note that for the purpose of this
paper we only consider weighted automata over the familiar field of
real numbers with standard addition and multiplication (see
\cite{Eilenberg1974,SalomaaSoittola1978,berstel2011noncommutative,KuichSalomaa1986,Mohri2009}
for more general definitions of WFAs over arbitrary semirings).
Functions mapping strings to real numbers can also be viewed as
non-commutative formal power series, which often helps deriving
rigorous proofs in formal language theory
\cite{SalomaaSoittola1978,berstel2011noncommutative,KuichSalomaa1986}. We
will not favor that point of view here, however, since we will not
need to make explicit mention of the algebraic properties offered by that perspective.

An alternative method to represent rational functions independently of
any WFA parametrization is via their \emph{Hankel matrices}.  The
Hankel matrix $\H_f \in \R^{\sstar \times \sstar}$ of a function
$f \colon \sstar \to \R$ is the infinite matrix with rows and
columns indexed by all strings with $\H_f(u, v) = f(u v)$ for all
$u, v \in \sstar$. By the theorem of Fliess \cite{Fliess1974} (see
also \cite{CarlylePaz1971} and \cite{berstel2011noncommutative}),
$\H_f$ has finite rank $n$ if and only if $f$ is rational and there
exists a WFA $A$ with $n$ states computing $f$, that is,
$\rank(f) = \rank(\H_f)$.

\subsection{Learning Scenario}

Let $\cZ$ denote a measurable subset of $\R$.  We assume a standard
supervised learning scenario where training and test points are drawn
i.i.d.\ according to some unknown distribution $D$ over
$\sstar \times \R$.

Let $\cF$ be a subset of the family of functions mapping from $\cX$ to
$\cY$, with $\cY \subseteq \R$, and let
$\ell \colon \cY \times \cZ \to \R_+$ be a loss function measuring the
divergence between the prediction $y \in \cY$ made by a function in
$\cF$ and the target label $z \in \cZ$. The learner's objective
consists of using a labeled training sample
$S = ((x_1, z_1), \ldots, (x_m, z_m))$ of size $m$ to select a
function $f \in \cF$ with small expected loss, that is
\begin{equation*}
\fL_D(f) = \E_{(x,z) \sim D}[\ell(f(x),z)] \enspace.
\end{equation*}
Our objective is to derive learning guarantees for broad families of weighted automata or rational functions used as hypothesis sets in learning algorithms. To do so, we will derive upper bounds on the Rademacher complexity of different classes of rational functions $f \colon \sstar \to \R$. Thus, we start with a brief introduction to the main definitions and results regarding the Rademacher complexity of an arbitrary class of functions $\cF = \{ f \colon \cX \to \cY \}$ where $\cX$ is the input space and $\cY \subseteq \R$ the output space. Let $D$ be a probability distribution over $\cX \times \cZ$ for some $\cZ \subseteq \R$ and denote by $D_{\cX}$ the marginal distribution over $\cX$. Suppose $S = (x_1, \ldots, x_m) \stackrel{\text{iid}}{\sim} D_{\cX}^m$ is a sample of $m$ i.i.d.\ examples drawn from $D$.  The \emph{empirical Rademacher  complexity} of $\cF$ on $S$ is defined as follows:
\begin{equation*}
\hfR_S(\cF) = \E\left[\sup_{f \in \cF} \frac{1}{m} \sum_{i=1}^m
\sigma_i f(x_i) \right] \enspace,
\end{equation*}
where the expectation is taken over the $m$ independent Rademacher random variables $\sigma_i \sim \dUnif(\{+1, -1\})$. The \emph{Rademacher complexity} of $\cF$ is defined as the expectation of $\hfR_S(\cF)$ over the draw of a sample $S$ of size $m$:
\begin{equation*}
\fR_m(\cF) = \E_{S \sim D_{\cX}^m} \left[\hfR_S(\cF)\right] \enspace.
\end{equation*}
The Rademacher complexity of a hypothesis class can be used to derive
generalization bounds for a variety of learning tasks
\cite{koltchinskii2000rademacher,bartlett2001rademacher,mohri2012foundations}. To
do so, we need to bound the Rademacher complexity of the associated
loss class, for a given loss function
$\ell \colon \cY \times \cZ \to \R_+$.

For a given hypothesis class $\cF$ the corresponding loss class
$\ell \circ \cF$ is given by the set of all functions
$\ell \circ f \colon \cX \times \cZ \to \R_+$ of the form
$(x,z) \mapsto \ell(f(x),z)$. By Talagrand's contraction lemma
\cite{ledtal}, the empirical Rademacher complexity
of $\ell \circ \cF$ can be bounded in terms of $\hfR_S(\cF)$, when
$\ell$ is $\mu$-Lipschitz with respect to its first argument for some
$\mu > 0$, that is when
\begin{equation*}
|\ell(y,z) - \ell(y',z)| \leq \mu |y -y'|
\end{equation*}
for all $y, y' \in \cY$ and $z \in \cZ$. In that case, the following
inequality holds: $\hfR_{S'}(\ell \circ \cF) \leq \mu \hfR_S(\cF)$,
where $S' = ((x_1,z_1),\ldots,(x_m,z_m))$ is a sample of size $m$ with
$(x_i, z_i) \in \cX \times \cZ$ and $S = (x_1, \ldots, x_m)$ denotes
the sample of elements in $\cX$ obtained from $S'$. When taking
expectations over $S' \stackrel{\text{iid}}{\sim} D^m$ and
$S \stackrel{\text{iid}}{\sim} D_{\cX}^m$ we obtain the same bound for
the Rademacher complexities
$\fR_m(\ell \circ \cF) \leq \mu \fR_m(\cF)$.  A typical example of a
loss function that is $\mu$-Lipschitz with respect to its first
argument is the absolute loss $\ell(y,z) = |y - z|$, which satisfies
the condition with $\mu = 1$ for $\cY = \cZ = \R$.

\section{Classes of Rational Functions}
\label{sec:classes}

In this section we introduce several classes of rational functions. Each of these classes is defined in terms of a different way to measure the complexity of rational functions. The first one is based on the weights of an explicit WFA representation, while the other two are based on intrinsic quantities associated to the function: the norm of the function, and the norm of the corresponding Hankel matrix when viewed as a linear operator on a certain Hilbert space. These three points of view measure different aspects of the complexity of a rational function, and each of them provides distinct benefits in the analysis of learning with WFAs. The Rademacher complexity of each of these classes will be analyzed in Sections~\ref{sec:Anr}, \ref{sec:Rpr}, and \ref{sec:Hpr}.

%Since the last two measures of complexity are not always finite, we also define variations of these measures that only look at the behavior of the function on a finite number of strings. 

\subsection{The Class $\cAnr$}

We start by considering the case where each rational function is given
by a fixed WFA representation. Our learning bounds would then
naturally depend on the number of states and the weights of the WFA
representations.

Fix an integer $n > 0$ and let $\cA_n$ denote the set of all WFAs with
$n$ states.  Note that any $A \in \cA_n$ is identified by the
$d = n (k n + 2)$ parameters required to specify its initial, final,
and transition weights.  Thus, we can identify $\cA_n$ with the vector
space $\R^d$ by suitably defining addition and scalar multiplication.
In particular, given $A, A' \in \cA_n$ and $c \in \R$, we define:
\begin{align*}
A + A' & = \wa + \wap = \langle \azero + \azero', \ainf + \ainf', \{\A_a + \A_a'\}
\rangle\\
c A & = c \wa = \langle c \azero, c \ainf, \{c \A_a\} \rangle \enspace.
\end{align*}
We can view $\cA_n$ as a normed vector space by endowing it with any
norm from the following family.  Let $p,q \in [1,+\infty]$ be H\"older
conjugates, i.e.\ $p^{-1} + q^{-1} = 1$.  It is easy to check that the
following defines a norm on $\cA_n$:
\begin{equation*}
  \norm{A}_{p,q} = \max\left\{\norm{\azero}_p,\norm{\ainf}_q,\max_a
    \norm{\A_a}_q\right\} \enspace,
\end{equation*}
where $\norm{\A}_q$ denotes the matrix norm induced by the
corresponding vector norm, that is
$\norm{\A}_q = \sup_{\norm{\v}_q = 1} \norm{\A \v}_q$.  Given
$p \in [1, +\infty]$ and $q = 1/(1 - 1/p)$, we denote by $\cAnr$ the
set of all WFAs $A$ with $n$ states and $\norm{A}_{p, q} \leq r$.
Thus, $\cAnr$ is the ball of radius $r$ at the origin in the normed
vector space $(\cA_n,\norm{\cdot}_{p, q})$.

\subsubsection{Examples}
We consider first the class of \emph{deterministic finite automata} (DFA). A DFA can be represented by a WFA where: $\azero$ is the indicator vector of the initial state; the entries of $\ainf$ are values in $\{0,1\}$ indicating whether a state is accepting or rejecting; and, for any $a \in \Sigma$ and any $i \in [n]$ we have that the $i$th row of $A_a$ is either the all zero vector if there is no transition from the $i$th state labeled by $a$, or an indicator vector with a one on the $j$th position if taking an $a$-transition from state $i$ leads to state $j$. Therefore, a DFA $A = \wa$ satisfies $\norm{A}_{1,\infty} \leq 1$ and $\cA_{n,1,1}$ contains all DFA with $n$ states.

Another important class of WFA contained in $\cA_{n,1,1}$ is that of \emph{probabilistic finite automata} (PFA). To represent a PFA as a WFA we consider automata where: $\azero$ is a probability distribution over possible initial states; the vector $\ainf$ contains stopping probabilities for every state; and for every $a \in \Sigma$ and $i, j \in [n]$ the entry $\A_a(i,j)$ represents the probability of transitioning from state $i$ to state $j$ while outputting the symbol $a$. Any WFA satisfying these constraints clearly has $\norm{\azero}_1 = 1$, $\norm{\ainf}_\infty \leq 1$, and $\norm{\A_a}_\infty = \max_i \sum_j |\A_a(i, j)| \leq 1$. The function $f_A$ computed by a PFA $A$ defines a probability distribution over $\sstar$; i.e.\ we have $f_A(x) \geq 0$ for all $x \in \sstar$ and $\sum_{x \in \sstar} f_A(x) = 1$.

\subsection{The Class $\cRpr$}\label{sec:defRpr}

Next, we consider an alternative quantity measuring the complexity of
rational functions that is \emph{independent} of any WFA
representation: their norm. Given $p \in [1, \infty]$ and $f \colon \sstar \to \R$ we use
$\norm{f}_p$ to denote the $p$-norm of $f$ given by
\begin{equation*}
\norm{f}_p = \bigg[ \sum_{x \in \sstar} |f(x)|^p \bigg]^{\frac 1 p} \enspace,
\end{equation*}
which in the case $p = \infty$ amounts to $\norm{f}_{\infty} = \sup_{x \in
\sstar} |f(x)|$.

Let $\cR_p$ denote the class of rational functions with finite
$p$-norm: $f \in \cR_p$ if and only if $f$ is rational and
$\norm{f}_p < +\infty$.  Given some $r > 0$ we also define $\cRpr$,
the class of functions with $p$-norm bounded by $r$:
\begin{equation*}
\cRpr = \left\{ f \colon \sstar \to \R \mid f \text{ rational and }
\norm{f}_p \leq r \right\} \enspace.
\end{equation*}
Note that this definition is independent of the WFA used to represent $f$.

%When learning from a distribution $D$ on $\sstar \times \R$ whose marginal distribution over $\sstar$ has finite support, it is not necessary to look at the norm of a function $f \colon \sstar \to \R$ over the whole of $\sstar$ in order to measure its complexity. For this reason we introduce a new set of semi-norms\footnote{Recall that a semi-norm satisfies the same axioms as a norm with the exception that it can assign the value zero to non-zero elements.} that only take into account the values of the function $f$ on a fixed set of strings. Given $W \subset \sstar$ and $p \in [1,\infty]$, we define the semi-norm
%\begin{equation*}
%\norm{f}_{p,W} = \bigg[ \sum_{x \in W} |f(x)|^p \bigg]^{\frac 1 p} \enspace,
%\end{equation*}
%which in the case $p = \infty$ amounts to $\norm{f}_{\infty,W} = \sup_{x \in W} |f(x)|$. Note that for any $f \colon \sstar \to \R$ we have $\norm{f}_{p,W} \leq \norm{f}_p$ for every $p$ and $W$, and in particular if $W$ is finite then $\norm{f}_{p,W}$ will always be finite even when $\norm{f}_p$ is not. On the other hand, if $f(x) = 0$ for all $x \in W$ then $\norm{f}_{p,W} = 0$ even if $f$ is not zero everywhere in $\sstar$, explaining why $\norm{\cdot}_{p,W}$ is only a semi-norm. Using these semi-norms, we define the classes of rational functions $\cRprW$ in the obvious way. It follows from our discussion that $\cR_{p,r} \subseteq \cRprW$ for every $W \subseteq \sstar$.

\subsubsection{Examples and Membership Testing}

If $A$ is a PFA, then the function $f_A$ is a probability distribution and we have $f_A \in \cR_{1,1}$ and by extension $\cR_{p,1}$ for all $p \in [1,+\infty]$. On the other hand, if $A$ is a DFA such that $f_A(x) = 1$ for infinitely many $x \in \sstar$, then $f_A \in \cR_{\infty,1}$, but $f_A \notin \cR_{p}$ for any $p < + \infty$. In fact, it is easy to see that for any $n \geq 0$ we have $\cA_{n,1,1} \subseteq \cR_{\infty}$. These examples show that $\cA_{n,1,1} \cap \cR_{1} \neq \emptyset$ and $\cA_{n,1,1} \cap (\cR_{\infty} \setminus \cR_{1}) \neq \emptyset$. Thus, the classes $\cR_p$ yield a more fine grained characterization of the complexity of rational functions than what the classes $\cA_{n,p,r}$ can provide in general.

However, while testing membership of a WFA in $\cA_{n,p,r}$ is a straightforward task, testing membership in any of the $\cR_p$ can be challenging. Membership in $\cR_{1,r}$ was shown to be semi-decidable in \cite{decideabsconv}. On the other hand, membership in $\cR_{2, r}$ can be decided in polynomial time \cite{cortes2007lp}. The inclusion $\cA_{n,1,1} \subseteq \cR_{\infty}$ gives an easy to test sufficient condition for membership in $\cR_{\infty}$.

%\borja{We should be able to say some more things here}

\subsection{The Class $\cHpr$}

Here, we introduce a third class of rational functions described via
their Hankel matrices, a quantity that is also independent of their
WFA representations.
To do so, we represent a function $f$ using its Hankel matrix $\H_f$,
interpret this matrix as a linear operator on a Hilbert space contained in the free vector space $\R^{\sstar}$, and consider the Schatten $p$-norm of $\H_f$ as a measure of complexity of $f$. To make this more precise we start by noting that the set
\begin{equation*}
\cL_2 = \left\{ f \colon \sstar \to \R \mid \norm{f}_2 < \infty \right\}
\end{equation*}
together with the inner product $\left< f,g \right> = \sum_{x \in \sstar} f(x) g(x)$ forms a separable Hilbert space. Note we have the obvious inclusion $\cR_2 \subset \cL_2$, but not all functions in $\cL_2$ are rational. Given an arbitrary function $f \colon \sstar \to \R$ we identify the Hankel matrix $\H_f$ with a (possibly unbounded) linear operator $\H_f \colon \cL_2 \to \cL_2$ defined by
\begin{equation*}
(\H_f g)(x) = \sum_{y \in \sstar} f(x y) g(y) \enspace.
\end{equation*}

Recall that an operator $\H_f$ is bounded when its operator norm is finite; i.e.\ $\norm{\H_f} = \sup_{\norm{g}_2 \leq 1} \norm{\H_f g}_2 < \infty$. Furthermore, a bounded operator is compact if it can be obtained as the limit of a sequence of bounded finite-rank operators under an adequate notion of convergence. In particular, bounded finite-rank operators are compact. Our interest in compact operators on Hilbert spaces stems from the fact that these are precisely the operators for which a notion equivalent to the SVD for finite matrices can be defined. Thus, if $f$ is a rational function of rank $n$ such that $\H_f$ is bounded (note this implies compactness by Fliess' theorem), then we can use the singular values $\sval_1 \geq \ldots \geq \sval_n$ of $\H_f$ as a measure of the complexity of $f$. The following result follows from \cite{bpp15} and gives a useful condition for the boundedness of $\H_f$.

\begin{lemma}
Suppose the function $f \colon \sstar \to \R$ is rational. Then $\H_f$ is bounded if and only if $\norm{f}_2 < \infty$.
\end{lemma}

We see that every Hankel matrix $\H_f$ with $f \in \cR_2$ has a well-defined SVD. Therefore, for any $f \in \cR_2$ it makes sense to define its Schatten--Hankel $p$-norm as the Schatten $p$-norm of its Hankel matrix: $\hknormp{f} = \normsch{\H_f} = \norm{(\sval_1,\ldots,\sval_n)}_p$, where $\sval_i = \sval_i(\H_f)$ is the $i$th singular value of $\H_f$ and
$\rank(\H_f) = n$. Using this notation, we can define several classes of rational functions. For a given $p \in [1, +\infty]$, we denote by $\cH_p$ the class of rational functions with $\hknormp{f} < \infty$ and, for any $r > 0$, we write $\cHpr$ the for class of rational functions with $\hknormp{f} \leq r$.

Note that the discussion above implies $\cH_p = \cR_2$ for every $p \in [1, +\infty]$, and therefore we can see the classes $\cHpr$ as providing an alternative stratification of $\cR_2$ than the classes $\cRtwor$. As a consequence of this containment we also have $\cR_1 \subset \cH_p$ for every $p$, and therefore the classes $\cH_p$ include all functions computed by probabilistic automata. Since membership in $\cR_2$ is efficiently testable \cite{cortes2007lp}, a polynomial time algorithm from \cite{bpp15} can be used to compute $\hknormp{f}$ and thus test membership in $\cHpr$.

\section{Rademacher Complexity of $\cAnr$}
\label{sec:Anr}

In this section, we present an upper bound on the Rademacher
complexity of the class of WFAs $\cAnr$.  To bound $\fR_m(\cAnr)$, we
will use an argument based on covering numbers.  We first introduce
some notation, then state our general bound and related corollaries,
and finally prove the main result of this section.

Let $S = (x_1, \ldots, x_m) \in \left(\sstar\right)^m$ be a sample
of $m$ strings with maximum length $L_S = \max_i |x_i|$.  The
expectation of this quantity over a sample of $m$ strings drawn
i.i.d.\ from some fixed distribution $D$ will be denoted by
$L_m = \E_{S \sim D^m}[L_S]$.  It is interesting at this point to note
that $L_m$ appears in our bound and introduces a dependency on the
distribution $D$ which will exhibit different growth rates depending
on the behavior of the tails of $D$.  For example, it is well known
that if the random variable $|x|$ for $x \sim D$ is
sub-Gaussian,\footnote{Recall that a non-negative random variable $X$
  is sub-Gaussian if $\P[X > k] \leq \exp(-\Omega(k^2))$,
  sub-exponential if $\P[X > k] \leq \exp(-\Omega(k))$, and follows a
  power-law with exponent $(s + 1)$ if
  $\P[X > k] \leq O(1/k^{s + 1})$.} then $L_m = O(\sqrt{\log m})$.
Similarly, if the tail of $D$ is sub-exponential, then
$L_m = O(\log m)$ and if the tail is a power-law with exponent
$s + 1$, $s > 0$, then $L_m = O(m^{1/s})$.  Note that in the latter
case the distribution of $|x|$ has finite variance if and only if
$s > 1$.

\begin{theorem}
\label{thm:RAnr}
The following inequality holds for every sample $S \in (\sstar)^m$:
\begin{equation*}
  \hfR_S(\cAnr) \leq \inf_{\eta > 0} \left(\eta +
    r^{L_S+2} \sqrt{\frac{2 n (k n+2) \log \Big(2 r + \frac{r^{L_S+2} (L_S+2)
        }{\eta}\Big)}{m}} \right) \enspace.
\end{equation*}
\end{theorem}
By considering the case $r = 1$ and choosing $\eta = (L_S + 2) / m$ we
obtain the following corollary.

\begin{corollary}
\label{cor:An1}
For any $m \geq 1$ and $n \geq 1$ the following inequalities holds:
\begin{align*}
\fR_m(\cA_{n,p,1}) &\leq \sqrt{\frac{2n(k n+2) \log(m+2)}{m}} + \frac{L_m + 2}{m}
\enspace, \\
\hfR_S(\cA_{n,p,1}) &\leq \sqrt{\frac{2n(k n+2) \log(m+2)}{m}} + \frac{L_S + 2}{m} \enspace.
%\label{eq:r1}
\end{align*}
\end{corollary}

\subsection{Proof of Theorem~\ref{thm:RAnr}}

We begin the proof by recalling several well-known facts and
definitions related to covering numbers (see e.g.\
\cite{devroye2001combinatorial}).  Let $V \subset \R^m$ be a set of
vectors and $S = (x_1, \ldots, x_m) \in (\sstar)^m$ a sample of size
$m$.  Given a WFA $A$, we define $A(S) \in \R^m$ by
$A(S) = (A(x_1), \ldots, A(x_m)) \in \R^m$.  We say that $V$ is an
$(\lone, \eta)$-cover for $S$ with respect to $\cAnr$ if for every
$A \in \cAnr$ there exists some $\v \in V$ such that
\begin{equation*}
\frac{1}{m} \norm{\v - A(S)}_1 = \frac{1}{m} \sum_{i = 1}^m |\v_i - A(x_i)| \leq
\eta \enspace.
\end{equation*}
The $\lone$-covering number of $S$ at level $\eta$ with respect to
$\cAnr$ is defined as follows:
\begin{equation*}
\cN_1(\eta,\cAnr,S) = \min\left\{ |V| \, \colon \, \text{$V \subset \R^m$ is an
$(\lone,\eta)$-cover for $S$ w.r.t.\ $\cAnr$} \right\} \enspace.
\end{equation*}
A typical analysis based on covering numbers would now proceed to
obtain a bound on the growth of $\cN_1(\eta,\cAnr,S)$ in terms of the
number of strings $m$ in $S$. Our analysis requires a slightly finer
approach where the size of $S$ is characterized by $m$ and $L_S$.
Thus, we also define for every integer $L \geq 0$ the following
covering number
\begin{equation*}
\cN_1(\eta,\cAnr,m,L) 
= \max_{S \in (\Sigma^{\leq L})^m} \cN_1(\eta,\cAnr,S) \enspace.
\end{equation*}

The first step in the proof of Theorem~\ref{thm:RAnr} is to bound
$\cN_1(\eta,\cAnr,m,L)$. In order to derive such a bound, we will make
use of the following technical results.

\begin{lemma}[Corollary 4.3 in \cite{vershynin-gfa}]
\label{lem:ball}
A ball of radius $R > 0$ in a real $d$-dimensional Banach space can be
covered by $R^d (2 + 1/\rho)^d$ balls of radius $\rho > 0$.
\end{lemma}

\begin{lemma}
\label{lem:Ax_diffAx}
Let $A, B \in \cAnr$.
Then the following hold for any $x \in \sstar$:
\begin{enumerate}
\item $|A(x)| \leq r^{|x|+2}$ \enspace,
\item $|A(x) - B(x)| \leq r^{|x|+1} (|x|+2) \norm{A-B}_{p,q}$ \enspace.
\end{enumerate}
\end{lemma}
\begin{proof}
  The first bound follows from applying H\"older's inequality and the
  sub-multiplicativity of the norms in the definition of
  $\norm{A}_{p,q}$ to \eqref{eqn:fAx}.  The second bound was proven in
  \cite{balle2012spectral}.
\end{proof}
Combining these lemmas yields the following bound on the covering
number $\cN_1(\eta,\cAnr,m,L)$.

\begin{lemma}
\label{lem:boundCN1}
\begin{equation*}
\cN_1(\eta,\cAnr,m,L) \leq r^{n(k n+2)}  \left(2 + \frac{r^{L+1} (L+2)
}{\eta}\right)^{n(k n+2)} \enspace.
\end{equation*}
\end{lemma}
\begin{proof}
  Let $d = n(k n+2)$.  By Lemma~\ref{lem:ball} and
  Lemma~\ref{lem:Ax_diffAx}, for any $\rho> 0$, there exists a finite
  set $\cC_\rho \subset \cAnr$ with
  $|\cC_\rho| \leq r^d (2 + 1/\rho)^d$ such that: for every
  $A \in \cAnr$ there exists $B \in \cC_\rho$ satisfying
  $|A(x) - B(x)| \leq r^{|x|+1} (|x|+2) \rho$ for every
  $x \in \sstar$.  Thus, taking $\rho = \eta / (r^{L+1} (L+2))$ we
  see that for every $S \in (\Sigma^{\leq L})^m$ the set
  $V = \{B(S) \colon B \in \cC_\rho \} \subset \R^m$ is an
  $\eta$-cover for $S$ with respect to $\cAnr$. 
\end{proof}

The last step of the proof relies on the following well-known result due to
Massart.

\begin{lemma}[Massart \cite{Massart2000}]
Given a finite set of vectors $V = \left\{\v_1,\ldots, \v_N\right\} \subset
\R^m$, the following holds
\begin{equation*}
\frac{1}{m} \E\left[\max_{\v \in V} \langle \boldsymbol{\sigma},\v
\rangle\right] \leq
\left(\max_{\v \in V} \norm{\v}_2\right) \frac{\sqrt{2 \log(N)}}{m} \enspace,
\end{equation*}
where the expectation is over the vector $\boldsymbol{\sigma} = (\sigma_1,
\ldots, \sigma_m)$ whose entries are independent Rademacher random variables
$\sigma_i \sim \dUnif(\{+1, -1\})$.
\end{lemma}

Fix $\eta > 0$ and let $V_{S,\eta}$ be an
$(\lone,\eta)$-cover for $S$ with respect to $\cAnr$.  By Massart's
lemma, we can write
\begin{equation}\label{eqn:massart}
\hfR_S(\cAnr) \leq \eta + \left(\max_{\v \in V_{S,\eta}} \norm{\v}_2 \right)
\frac{\sqrt{2 \log |V_{S,\eta}|}}{m} \enspace.
\end{equation}
Since $|A(x_i)| \leq r^{L_S + 2}$ by Lemma~\ref{lem:Ax_diffAx}, we can
restrict the search for $(\lone,\eta)$-covers for $S$ to sets
$V_{S,\eta} \subset \R^m$ where all $\v \in V_{S,\eta}$ must satisfy
$\norm{\v}_\infty \leq r^{L_S + 2}$.  By construction, such a covering
satisfies
$\max_{\v \in V_{S,\eta}} \norm{\v}_2 \leq r^{L_S+2} \sqrt{m}$.
Finally, plugging in the bound for $|V_{S,\eta}|$ given by
Lemma~\ref{lem:boundCN1} into \eqref{eqn:massart} and taking the
infimum over all $\eta > 0$ yields the desired result.

\section{Rademacher Complexity of $\cRpr$}
\label{sec:Rpr}

In this section, we study the complexity of rational functions from a
different perspective. Instead of analyzing their complexity in terms
of the parameters of WFAs computing them, we consider an intrinsic associated
quantity: their norm.  We present upper bounds on the
Rademacher complexity of the classes of rational functions $\cRpr$ for
any $p \in [1, +\infty]$ and $r > 0$.

It will be convenient for our analysis to identify a rational function
$f \in \cRpr$ with an infinite-dimensional vector
$\vf \in \R^{\sstar}$ with $\norm{\vf}_p \leq r$. That is, $\vf$ is
an infinite vector indexed by strings in $\sstar$ whose $x$th entry
is $\vf_x = f(x)$.  An important observation is that using this
notation, for any given $x \in \sstar$, we can write $f(x)$ as the
inner product $\langle \vf, \ve_x \rangle$, where
$\ve_x \in \R^{\sstar}$ is the indicator vector corresponding to
string $x$.

\begin{theorem}
\label{thm:Rpr}
Let $p^{-1} + q^{-1} = 1$. Let $S = (x_1, \ldots, x_m)$ be a sample of
$m$ strings.  Then, the following holds for any $r > 0$:
\begin{equation*}
\hfR_S(\cRpr) = \frac{r}{m} \E\left[ \bigg\| \sum_{i = 1}^m \sigma_i
  \ve_{x_i} \bigg\|_q
\right] \enspace,
\end{equation*}
where the expectation is over the $m$ independent Rademacher random
variables $\sigma_i \sim \dUnif(\{+1,-1\})$.
\end{theorem}
\begin{proof}
  In view of the notation just introduced described, we can write
\begin{align*}
\hfR_S(\cRpr) 
% & = \E\left[\sup_{f \in \cRpr} \frac{1}{m} \sum_{i = 1}^m \sigma_i f(x_i) \right]
= \E\left[\sup_{f \in \cRpr} \frac{1}{m} \sum_{i = 1}^m \langle \vf, \sigma_i
\ve_{x_i} \rangle \right] 
& = \frac{1}{m} \E\left[\sup_{f \in \cRpr} \bigg\langle \vf, \sum_{i = 1}^m \sigma_i
\ve_{x_i} \bigg\rangle \right] \\
& = \frac{r}{m} \E\left[\bigg\| \sum_{i = 1}^m \sigma_i \ve_{x_i} \bigg\|_q \right] \enspace,
\end{align*}
where the last inequality holds by definition of the dual norm.
%\ignore{
%Now suppose the Rademacher random variables $\mat{\sigma} = (\sigma_i)_{i = 1}^m$
%are fixed and define the function $g \colon \sstar \to \R$ given by
%\begin{equation*}
%g(x) = \langle \ve_{x}, \sum_{i = 1}^m \sigma_i \ve_{x_i} \rangle = \sum_{i = 1}^m
%\sigma_i \one_{x = x_i} \enspace.
%\end{equation*}
%Note that $g$ depends on $S$ and the values of the $\sigma_i$.  Next
%we use $g$ to define another function $f_r \colon \sstar \to \R$
%given by
%\begin{equation*}
%f_r(x) = 
%\begin{cases}
%r \norm{g}_q^{1-q} \frac{|g(x)|^q}{g(x)} & \text{if $g(x) \neq 0$} \enspace,\\
%0 & \text{if $g(x) = 0$} \enspace.
%\end{cases}
%\end{equation*}
%An easy calculation shows that by construction we have
%$\norm{f_r}_p = r$.  Observe also that $f_r$ has finite support; i.e.\
%there exists a finite number of strings $x \in \sstar$ such that
%$f_r(x) \neq 0$. Since this last observation implies the Hankel matrix
%of $f_r$ will have finite rank, we see that $f_r$ is rational and
%therefore $f_r \in \cRpr$. Furthermore, another calculation shows that
%we actually have $\langle f_r , g \rangle = r \norm{g}_q$. This last
%fact means that $f_r$ attains the maximum of H\"older's inequality
%$\langle f, g \rangle \leq \norm{f}_p \norm{g}_q$ among all
%$f \in \cRpr$. Thus, we can conclude that
%\begin{equation*}
%\E\left[\sup_{f \in \cRpr} \bigg\langle \vf, \sum_{i = 1}^m \sigma_i
%\ve_{x_i} \bigg\rangle \right] = 
%r\E\left[\bigg\| \sum_{i = 1}^m \sigma_i \ve_{x_i} \bigg\|_q
%\right] \enspace,
%\end{equation*}
%which yields the desired bound.
%\qed}
\end{proof}
The next corollaries give non-trivial bounds on the Rademacher
complexity in the case $p = 1$ and the case $p = 2$.
\begin{corollary}
\label{cor:R2r}
For any $m \geq 1$ and any $r > 0$, the following inequalities hold:
\begin{equation*}
\frac{r}{\sqrt{2 m}} \leq \fR_m(\cRtwor) \leq \frac{r}{\sqrt{m}}.
\end{equation*}
\end{corollary}
\begin{proof}
  The upper bound follows directly from Theorem~\ref{thm:Rpr} and
  Jensen's inequality:
\begin{equation*}
\E\left[ \bigg\| \sum_{i = 1}^m \sigma_i \ve_{x_i} \bigg\|_2 \right] 
\leq \sqrt{\E\left[ \bigg\| \sum_{i = 1}^m \sigma_i \ve_{x_i} \bigg\|_2^2 \right]}
= \sqrt{m} \enspace.
\end{equation*}
The lower bound is obtained using Khintchine--Kahane's inequality (see
appendix of \cite{mohri2012foundations}):
\begin{equation*}
\E\left[ \bigg\| \sum_{i = 1}^m \sigma_i \ve_{x_i} \bigg\|_2 \right]^2
\geq \frac{1}{2} \E\left[ \bigg\| \sum_{i = 1}^m \sigma_i \ve_{x_i}
  \bigg\|^2_2 \right]
= \frac{m}{2},
\end{equation*}
which completes the proof.
\end{proof}
The following definition will be needed to present our next corollary. Given a sample $S = (x_1, \ldots, x_m)$ and a string $x \in \sstar$ we denote by $s_x = |\{ i \colon x_i = x \}|$ the number of times $x$ appears in $S$.  Let $C_S = \max_{s \in \sstar} s_x$ and note we have the straightforward bounds $1 \leq C_S \leq m$.

\begin{corollary}
\label{cor:R1r}
For any $m \geq 1$, any $S \in (\sstar)^m$, and any $r > 0$, the following upper bound holds:
\begin{equation*}
\hfR_S(\cRoner) \leq \frac{r \sqrt{2 C_S \log(2 m)}}{m}.
\end{equation*}
\end{corollary}
\begin{proof}
Let $S = (x_1,\ldots,x_m)$ be a sample with $m$ strings. For any $x \in \sstar$
define the vector $\v_x \in \R^m$ given by $\v_x(i) = \one_{x_i = x}$. Let $V$
be the set of vectors $\v_x$ which are not identically zero, and note we have
$|V| \leq m$. Also note that by construction we have $\max_{\v_x \in V}
\norm{\v_x}_2 = \sqrt{C_S}$.
Now, by Theorem~\ref{thm:Rpr} we have
\begin{equation*}
\hfR_S(\cRoner) = \frac{r}{m} \E\left[ \bigg\| \sum_{i = 1}^m \sigma_i
\ve_{x_i} \bigg\|_\infty \right] =
\frac{r}{m} \E\left[ \max_{\v_x \in V \cup (-V)} \langle \boldsymbol{\sigma}, \v_x\rangle
\right] \enspace.
\end{equation*}
Therefore, using Massart's Lemma we get
\begin{equation*}
\hfR_S(\cRoner) \leq \frac{r \sqrt{2 C_S \log (2m)}}{m} \enspace.
\qedhere
\end{equation*}
%The result now follows from taking the expectation over $S$ and using Jensen's
%inequality to see that $\E[\sqrt{C_S}] \leq \sqrt{C_m}$.
\end{proof}

Note in this case we cannot rely on the Khintchine--Kahane inequality to obtain lower bounds because there is no version of this inequality for the case $q = \infty$.

We can easily convert the above empirical bound into a standard Rademacher complexity bound by defining the expectation $C_m = \E_{S \sim D^m}[C_S]$ over a distribution $D$ on $\sstar$. Note that $C_m$ is the expected maximum number of collisions (repeated strings) in a sample of size $m$ drawn from $D$. We shall provide a bound for $C_m$ in terms of $m$ in Section~\ref{sec:distparams}.

%\subsection{$\cRpr$ Restricted to a Subset}
%
%\borja{TODO: Put the result and proof here.}

\section{Rademacher Complexity of $\cHpr$}
\label{sec:Hpr}

In this section, we present our last set of upper bounds on the
Rademacher complexity of WFAs. Here, we characterize the complexity of
WFAs in terms of the spectral properties of their Hankel matrix.

The Hankel matrix of a function $f\colon \sstar \to \R$ is the
bi-infinite matrix $\H_f \in \R^{\sstar \times \sstar}$ whose
entries are defined by $\H_f(u, v) = f( u v )$. Note that any string
$x \in \sstar$ admits $|x| + 1$ decompositions $x = u v$ into a
prefix $u \in \sstar$ and a suffix $v \in \sstar$. Thus, $\H_f$
contains a high degree of redundancy: for any $x \in \sstar$, $f(x)$
is the value of at least $|x| + 1$ entries of $\H_f$ and we can write
$f(x) = \ve_u^\top \H_f \ve_v$ for any decomposition $x = u v$.

Let $\sval_i(\mat{M})$ denote the $i$th singular value of a matrix
$\mat{M}$.  For $1 \leq p \leq \infty$, let $\normsch{\mat{M}}$ denote
the $p$-Schatten norm of $\mat{M}$ defined by
$\normsch{\mat{M}} = \big[\sum_{i \geq 1} \sval_i(\mat{M})^p
\big]^{\frac 1 p}$.

\begin{theorem}
\label{thm:Hpr}
Let $p, q \geq 1$ with $p^{-1} + q^{-1} = 1$ and let
$S = (x_1, \ldots, x_m)$ be a sample of $m$ strings in $\sstar$.
For any decomposition $x_i = u_i v_i$ of the strings in $S$ and any
$r > 0$, the following inequality holds:
\begin{equation*}
\hfR_S(\cHpr)
\leq
\frac{r}{m} \E\left[ \bigg\| \sum_{i = 1}^m \sigma_i \ve_{u_i}
\ve_{v_i}^\top \bigg\|_{\mathrm{S}, q} \right]
\enspace.
\end{equation*}
\end{theorem}
\begin{proof}
  For any $1 \leq i \leq m$, let $x_i = u_i v_i$ be an arbitrary
  decomposition and let
  $\mR = \sum_{i = 1}^{m} \sigma_i \ve_{u_i} \ve_{v_i}^\top$. Then,
in view of the identity $f(x_i) = \ve_{u_i}^\top \H_f
\ve_{v_i} = \trace(\ve_{v_i} \ve_{u_i}^\top \H_f)$, we can use the linearity of the trace to write
\begin{align*}
\hfR_S(\cHpr) 
& =
\E\left[\sup_{f \in \cHpr} \frac{1}{m} \sum_{i = 1}^m \sigma_i \ve_{u_i}^\top \H_f
\ve_{v_i} \right]
\\ &=
\frac{1}{m} \E\left[\sup_{f \in \cHpr} \sum_{i = 1}^m \trace\left( \sigma_i
\ve_{v_i} \ve_{u_i}^\top \H_f \right) \right]
%\\ &=
=
\frac{1}{m} \E\left[\sup_{f \in \cHpr} \langle \mR, \H_f \rangle \right]
\enspace.
\end{align*}
Then, by von Neumann's trace inequality \cite{vonneumann} and
H\"older's inequality, the following holds:
\begin{align*}
\E\left[\sup_{f \in \cHpr} \langle \mR, \H_f \rangle \right]
&\leq
\E\left[\sup_{f \in \cHpr} \sum_{j \geq 1} \sval_j(\mR) \cdot \sval_j(\H_f) \right]
\\ &\leq
\E\left[\sup_{f \in \cHpr} \normschq{\mR} \normsch{\H_f} \right]
%\\ &=
=
r \E \big[\normschq{\mR} \big]
\enspace,
\end{align*}
which completes the proof.
\end{proof}
Note that, in this last result, the equality condition for von
Neumann's inequality cannot be used to obtain a lower bound on
$\hfR_S(\cHpr)$ since it requires the simultaneous diagonalizability
of the two matrices involved, which is difficult to control in the
case of Hankel matrices.

As in the previous sections, we now proceed to derive specialized
versions of the bound of Theorem~\ref{thm:Hpr} for the cases $p = 1$
and $p = 2$.  First, note that the corresponding $q$-Schatten norms
have given names: $\normschtwo{\mR} = \normfro{\mR}$ is the Frobenius
norm, and $\normschinf{\mR} = \normop{\mR}$ is the operator norm.

\begin{corollary}
\label{cor:H2r}
For any $m \geq 1$ and any $r > 0$, the Rademacher complexity of
$\cHtwor$ can be bounded as follows:
\begin{equation*}
\fR_m(\cHtwor) \leq \frac{r}{\sqrt{m}}.
\end{equation*}
\end{corollary}
\begin{proof}
In view of Theorem~\ref{thm:Hpr} and using Jensen's inequality, we
can write
\begin{align*}
\fR_m(\cHtwor)
\leq \frac{r}{m} \E \big[ \normfro{\mR} \big]
& \leq \frac{r}{m} \sqrt{\E \big[\| \mR \|_F^2 \big]}\\
& = \frac{r}{m} \sqrt{\E \bigg[\sum_{i, j = 1}^m \sigma_i \sigma_j
  \langle \ve_{u_i} \ve_{v_i}^\top, \ve_{u_j} \ve_{v_j}^\top\rangle \bigg]}\\
& = \frac{r}{m} \sqrt{\E \Big[\sum_{i = 1}^m \langle \ve_{u_i}
  \ve_{v_i}^\top, \ve_{u_i} \ve_{v_i}^\top\rangle \Big]}
= \frac{r}{\sqrt{m}} \enspace,
\end{align*}
which concludes the proof.
\end{proof}

To bound the Rademacher complexity of $\cHpr$ in the case $p = 1$ we will need the following moment bound for the operator norm of a random matrix from \cite{tropp2015introduction}.

\begin{theorem}[Corollary~7.3.2 \cite{tropp2015introduction}]
\label{thm:tropp}
Suppose $\mM = \sum_{i} \mM_i$ is a sum of i.i.d.\ random matrices
with $\E[\mM_i] = \mat{0}$ and $\normop{\mM_i} \leq M$. Let
$\sum_i \E[\mM_i \mM_i^\top] \preccurlyeq \mV_1$,
$\sum_i \E[\mM_i^\top \mM_i] \preccurlyeq \mV_2$, and
$\mV = \diag(\mV_1,\mV_2)$. If $d = \trace(\mV)/\normop{\mV}$ and
$\nu = \normop{\mV}$, then we have
\begin{equation*}
  \E[\normop{\mM}] 
  \leq \frac{2}{3}\left(1 + \frac{4}{\log 2}\right) M \log(d + 1) +
  \left(1 + \frac{4}{\sqrt{2 \log 2}}\right) \sqrt{2 \nu \log(d + 1) }  \enspace.
\end{equation*}
\end{theorem}

We now introduce a combinatorial number depending on $S$ and the
decomposition selected for each string $x_i$. Let
$U_S = \max_{u \in \sstar} |\{ i \colon u_i = u \}|$ and
$V_S = \max_{v \in \sstar} |\{ i \colon v_i = v \}|$. Then, we define
$W_S = \min \max\{U_S, V_S\}$, where then minimum is taken over all
possible decompositions of the strings in $S$. It is easy to show that
we have the bounds $1 \leq W_S \leq m$. Indeed, for the case $W_S = m$
consider a sample with $m$ copies of the empty string, and for the
case $W_S = 1$ consider a sample with $m$ different strings of length
$m$. The following result can be stated using this definition.

\begin{corollary}
\label{cor:H1r}
For any $m \geq 1$, any $S \in (\sstar)^m$, and any $r > 0$, the following upper bound holds:
\begin{equation*}
\hfR_S(\cHoner) \leq \frac{r}{m} 
\left[\frac{2}{3}\left(1 + \frac{4}{\log 2}\right) \log(2 m+1) +
  \left(1 + \frac{4}{\sqrt{2\log 2}}\right) \sqrt{2 W_S \log(2 m + 1) } \right]
\enspace.
\end{equation*}
\end{corollary}
\begin{proof}
First note that we can apply Theorem~\ref{thm:tropp} to the random matrix $\mR$ by letting $\mV_1 = \sum_i \ve_{u_i} \ve_{u_i}^\top$ and $\mV_2 = \sum_i \ve_{v_i} \ve_{v_i}^\top$. In this case we have $d = 2m$, $\nu = \max\{\normop{\sum_i \ve_{u_i} \ve_{u_i}^\top}, \normop{\sum_i \ve_{v_i} \ve_{v_i}^\top} \}$, and we get:
\begin{equation*}
\E[\normop{\mR}] \leq \left(\frac{2}{3} + \frac{8}{3 \log 2}\right) \log(2 m+1) + \left(\sqrt{2} + \frac{4}{\sqrt{\log 2}}\right) \sqrt{ \nu \log(2 m+1) }
\enspace.
\end{equation*}
Next, observe that $\mV_1 = \sum_i \ve_{u_i} \ve_{u_i}^\top \in \R^{\sstar \times \sstar}$ is a diagonal matrix with $\mV_1(u,u) = \sum_i \one_{u = u_i}$.  Thus, $\normop{\mV_1} = \max_u \mV_1(u,u) = \max_{u \in \sstar} |\{ i \colon u_i = u \}| = U_S$. Similarly, we have $\normop{\mV_2} = V_S$. Thus, since the decomposition of the strings in $S$ is arbitrary, we can choose it such that $\mu = W_S$.
Applying Theorem~\ref{thm:Hpr} now yields the desired bound.
\end{proof}

We can again convert the above empirical bound into a standard Rademacher complexity bound by defining the expectation $W_m = \E_{S \sim D^m}[W_S]$ over a distribution $D$ on $\sstar$. We provide a bound for $W_m$ in terms of $m$ in next section.

\section{Distribution-Dependent Rademacher Complexity Bounds}\label{sec:distparams}

The bounds for the Rademacher complexity of $\cR_{1,r}$ and $\cH_{1,r}$ we give above identify two important distribution-dependent parameters $C_m = \E_S[C_S]$ and $W_m = \E_S[W_S]$ that reflect the impact of the distribution $D$ on the complexity of learning these classes of rational functions. We now use upper bounds on $C_m$ and $W_m$ in terms of $m$ to give bounds for the Rademacher complexities $\fR_m(\cR_{1,r})$ and $\fR_m(\cH_{1,r})$.

We start by rewriting $C_S$ in a convenient way. Let $\cE = \{ e_x \colon \sstar \to \R | x \in \sstar \}$ be the class of all indicator on $\sstar$ given by $e_x(y) = 1$ if $x = y$ and $e_x(y) = 0$ otherwise. Recall that given $S = (x_1, \ldots, x_m)$ we defined $s_x = |\{i \colon x_i = x\}|$ and $C_S = \sup_{x \in \sstar} s_x$. Using $\cE$ we can rewrite these as $s_x = \sum_{i=1}^m e_x(x_i)$ and
\begin{equation*}
C_S = \sup_{e_x \in \cE} \sum_{i=1}^m e_x(x_i) \enspace.
\end{equation*}
Let $D_{\max} = \max_{x \in \sstar} \P_D[x]$ be the maximum probability of any strings with respect to the distribution $D$.

\begin{lemma}
\begin{equation*}
m D_{\max} \leq C_m \leq m D_{\max} + O(\sqrt{m}) \enspace.
\end{equation*}
\end{lemma}
\begin{proof}
We can bound $C_m = \E_S[C_S]$ as follows:
\begin{align*}
C_m &= \E_{S \sim D^m}\left[ \sup_{e_x \in \cE} \sum_{i=1}^m e_x(x_i) \right] \\
&= \E_{S \sim D^m}\left[ \sup_{e_x \in \cE} \sum_{i=1}^m \left(e_x(x_i) + \E_{x_i' \sim D}[e_x(x_i')] -  \E_{x_i' \sim D}[e_x(x_i')] \right) \right] \\
&\leq \E_{S \sim D^m}\left[ \sup_{e_x \in \cE} \sum_{i=1}^m \E_{x_i' \sim D}[e_x(x_i')]\right]
+ \E_{S \sim D^m}\left[ \sup_{e_x \in \cE} \sum_{i=1}^m \left(e_x(x_i) -  \E_{x_i' \sim D}[e_x(x_i')] \right) \right]
 \\
&= m \sup_{e_x \in \cE} \E_{x' \sim D}[e_x(x')]
+ \E_{S \sim D^m}\left[ \sup_{e_x \in \cE} \sum_{i=1}^m \left(e_x(x_i) -  \E_{x_i' \sim D}[e_x(x_i')] \right) \right]
 \\
&\leq m \sup_{e_x \in \cE} \E_{x' \sim D}[e_x(x')] + \E_{S \sim D^m} \left[ \sup_{e_x \in \cE} \left| \sum_{i=1}^m \left(e_x(x_i) - \E_{x_i' \sim D}[e_x(x_i')]\right) \right| \right] \enspace.
%&\leq m \sup_{x \in \sstar} \P_{x' \sim D}[x' = x] + 2 m \fR_m(\cE) \\
%&\leq m D_{\max} + O\left(\sqrt{m}\right) \enspace,
\end{align*}
Now note on the one hand we can write $\sup_{e_x \in \cE} \E_{x' \sim D}[e_x(x')] = \sup_{x \in \sstar} \P_{x' \sim D}[x' = x] = D_{\max}$. On the other hand, a standard symmetrization argument yields:
\begin{equation*}
\E_{S \sim D^m} \left[ \sup_{e_x \in \cE} \left| \sum_{i=1}^m \left(e_x(x_i) - \E_{x_i' \sim D}[e_x(x_i')]\right) \right| \right] \leq 2 m \fR_m(\cE) = O(\sqrt{m}) \enspace,
\end{equation*}
where in the last inequality we used that the VC-dimension of $\cE$ is $1$, in which case Dudley's chaining method \cite{dudley1999uniform} yields $\fR_m(\cE) \leq C \sqrt{1/m}$ for some universal constant $C > 0$. Note that by Jensen's inequality we also have
\begin{equation*}
m \sup_{e_x \in \cE} \E_{x' \sim D}[e_x(x')] = \sup_{e_x \in \cE} \E_{S \sim D^m}\left[ \sum_{i=1}^m e_x(x_i)\right] \leq \E_{S \sim D^m}\left[ \sup_{e_x \in \cE} \sum_{i=1}^m e_x(x_i)\right] \enspace,
\end{equation*}
and therefore the bound is tight up to the lower order terms.
\end{proof}

A straightforward application of Jensen's inequality now yields the following.

\begin{corollary}
\label{cor:R1rm}
For any $m \geq 1$ and any $r > 0$ we have:
\begin{equation*}
\fR_m(\cRoner) \leq \frac{r}{\sqrt{m}} \sqrt{2 (D_{\max} + O(\sqrt{1/m})) \log(2 m)}.
\end{equation*}
\end{corollary}

Next we provide bounds for $W_m$. Given a sample
$S = (x_1, \ldots, x_m)$ we will say that the tuples of pairs of
strings
$S' = ((u_1,v_1),\ldots, (u_m,v_m)) \in (\sstar \times \sstar)^m$ form
a \emph{split} of $S$ if $x_i = u_i v_i$ for all $1 \leq i \leq m$. We
denote by $S^\vee$ the set of all possible splits of a sample $S$. We
also define coordinate projections
$\pi_j \colon \sstar \times \sstar \to \sstar$ given by
$\pi_1(u,v) = u$ and $\pi_2(u,v) = v$. Now recall that
$W_m = \E_S[W_S]$ and note we can rewrite the definition of $W_S$ as
\begin{align*}
W_S &= \min_{S' \in S^\vee} \max_{j=1,2} \sup_{e_x \in \cE} \sum_{i=1}^m e_x(\pi_j(u_i,v_i)) \\
&= \min_{S' \in S^\vee} \sup_{e \in \cE^\vee} \sum_{i=1}^m e(u_i,v_i)
\enspace,
\end{align*}
where $\cE^\vee = (\cE \circ \pi_1) \cup (\cE \circ \pi_2)$ and $\cE \circ \pi_j$ is the set of functions of the form $e_x(\pi_j(u,v))$. Finally, given a distribution $D$ over $\sstar$ we define the parameter
\begin{equation*}
D_{\max}^\vee = \sup_{x \in \sstar} \max\left\{\sum_{v \in \sstar} \frac{1}{|x| + |v| + 1} \P_D[x v], \sum_{u \in \sstar} \frac{1}{|x| + |u| + 1} \P_D[u x] \right\} \enspace.
\end{equation*}
With these definitions we have the following result.

\begin{lemma}
\begin{equation*}
W_m \leq m D_{\max}^\vee + O(\sqrt{m}) \enspace.
\end{equation*}
\end{lemma}
\begin{proof}
We start by upper bounding the $\min_{S' \in S^\vee}$ with the expectation $\E_{S' \sim \dUnif(S^\vee)}$ over a split chosen uniformly at random:
\begin{align*}
W_m &= \E_{S \sim D^m}\left[ \min_{S' \in S^\vee} \sup_{e \in \cE^\vee} \sum_{i=1}^m e(u_i,v_i) \right] \\
&\leq \E_{S \sim D^m} \E_{S' \sim \dUnif(S^\vee)} \left[\sup_{e \in \cE^\vee} \sum_{i=1}^m e(u_i,v_i) \right] \\
&\leq \sup_{e \in \cE^\vee} \E_{S \sim D^m} \E_{S' \sim \dUnif(S^\vee)} \left[ \sum_{i=1}^m e(u_i,v_i) \right] \\
&+ \E_{S \sim D^m} \E_{S' \sim \dUnif(S^\vee)}  \left[\sup_{e \in \cE^\vee} \left|\sum_{i=1}^m \left(e(u_i,v_i) - \E_{x_i' \sim D} \E_{(u_i',v_i') \sim \dUnif(\{x_i'\}^\vee)}[e(u_i',v_i')] \right)\right|\right] \enspace.
\end{align*}
The same standard argument we used above shows that the second term in the last sum above can be bounded by $2 m \fR_m(\cE^\vee) = O(\sqrt{m})$. To compute the first term in the sum note that given a string $y$ and a random split $(u,v) \sim \dUnif(\{y\}^\vee)$, the probability that $u = x$ for some fixed $x \in \sstar$ is $1/(|y|+1)$ if $x$ is a prefix of $y$ and $0$ otherwise. Thus, we let $e = e_x \circ \pi_1 \in \cE^\vee$ and write
\begin{align*}
\E_{S \sim D^m} \E_{S' \sim \dUnif(S^\vee)} \left[ \sum_{i=1}^m e(u_i,v_i) \right]
&=
m \E_{x' \sim D} \E_{(u,v) \sim \dUnif(\{x'\}^\vee)} e_x(u) \\
&=
m \P_{x' \sim D, (u,v) \sim \dUnif(\{x'\}^\vee)}[u = x] \\
&=
m \sum_{x' \in x \sstar} \frac{1}{|x'|+1} \P_D[x'] \\
&=
m \sum_{v \in \sstar} \frac{1}{|x| + |v| + 1} \P_D[x v] \enspace.
\end{align*}
Similarly, if we have $e = e_x \circ \pi_2 \in \cE^\vee$ then
\begin{equation*}
\E_{S \sim D^m} \E_{S' \sim \dUnif(S^\vee)} \left[ \sum_{i=1}^m e(u_i,v_i) \right]
=
m \sum_{u \in \sstar} \frac{1}{|x| + |u| + 1} \P_D[u x] \enspace.
\end{equation*}
Thus, we can combine these equations to show that $W_m \leq m D_{\max}^\vee + O(\sqrt{m})$.
\end{proof}

Using Jensen's inequality we now obtain the following bound.

\begin{corollary}
\label{cor:H1rm}
For any $m \geq 1$ and any $r > 0$ we have:
\begin{align*}
\fR_m(\cHoner) &\leq 
\left(\frac{2}{3} + \frac{8}{3 \log 2}\right) \frac{r \log(2 m+1)}{m} \\
&+ \left(\sqrt{2} + \frac{4}{\sqrt{\log 2}}\right) \frac{r}{\sqrt{m}} \sqrt{ (D_{\max}^\vee + O(\sqrt{1/m})) \log(2 m+1)}
\enspace.
\end{align*}
\end{corollary}

\section{Learning and Sample Complexity Bounds}\label{sec:bounds}

We now have all the ingredients to give generalization bounds for learning with weighted automata. In particular, we will give bounds for learning with a Lipschitz bounded loss function on all the classes of weighted automata and rational functions considered above. In cases where we have different bounds for the empirical and expected Rademacher complexities we also give two versions of the bound. All these bounds can be used to derive learning algorithms for weighted automata provided the right-hand side can be optimized over the corresponding hypothesis class. We will discuss in the next section what are the open problems related to obtaining efficient algorithms to solve these optimization problems. The proofs of these theorems are a straightforward combination of the bounds on the Rademacher complexity with well-known generalization bounds \cite{mohri2012foundations}.

\begin{theorem}
  Let $D$ be a probability distribution over $\sstar \times \R$ and
  let $S = ((x_i,y_i))_{i=1}^m$ be a sample of $m$ i.i.d.\ examples
  from $D$. Assume that the loss $\ell \colon \R \times \R \to \R_+$
  is $M$-bounded and $\mu$-Lipschitz with respect to its first
  argument.  Fix $\delta > 0$. Then, the following holds:
\begin{enumerate}
\item For all $n \geq 1$ and $p \in [1,+\infty]$, with probability at least $1 - \delta$ the following holds simultaneously for all $A \in \cA_{n,p,1}$:
\begin{equation*}
\fL_D(A) \leq \hfL_S(A) + \sqrt{\frac{8 \mu^2 n(k n+2) \log(m+2)}{m}} + \frac{2 \mu (L_m + 2) }{m} + M \sqrt{\frac{\log (1/\delta)}{2 m}} \enspace.
\end{equation*}
\item For all $r > 0$, with probability at least $1 - \delta$ the following holds simultaneously for all $f \in \cR_{2,r}$:
\begin{equation*}
\fL_D(f) \leq \hfL_S(f) + \frac{2 \mu r}{\sqrt{m}} + M \sqrt{\frac{\log (1/\delta)}{2 m}} \enspace.
\end{equation*}
\item For all $r > 0$, with probability at least $1 - \delta$ the following holds simultaneously for all $f \in \cR_{1,r}$:
\begin{equation*}
\fL_D(f) \leq \hfL_S(f) + 
\frac{2 \mu r}{\sqrt{m}} \sqrt{2 (D_{\max} + O(\sqrt{1/m})) \log(2 m)}
+ M \sqrt{\frac{\log (1/\delta)}{2 m}} \enspace.
\end{equation*}
\item For all $r > 0$, with probability at least $1 - \delta$ the following holds simultaneously for all $f \in \cH_{2,r}$:
\begin{equation*}
\fL_D(f) \leq \hfL_S(f) + \frac{2 \mu r}{\sqrt{m}} + M \sqrt{\frac{\log (1/\delta)}{2 m}} \enspace.
\end{equation*}
\item For all $r > 0$, with probability at least $1 - \delta$ the following holds simultaneously for all $f \in \cH_{1,r}$:
\begin{align*}
\fL_D(f) \leq \hfL_S(f) &+ \left(\sqrt{2} + \frac{4}{\sqrt{\log 2}}\right) \frac{2 \mu r}{\sqrt{m}} \sqrt{ (D_{\max}^\vee + O(\sqrt{1/m})) \log(2 m+1)} \\
&+\left(\frac{2}{3} + \frac{8}{3 \log 2}\right) \frac{2 \mu r \log(2 m+1)}{m}
+ M \sqrt{\frac{\log (1/\delta)}{2 m}} \enspace.
\end{align*}
\end{enumerate}
\end{theorem}

\begin{theorem}
  Let $D$ be a probability distribution over $\sstar \times \R$ and
  let $S = ((x_i,y_i))_{i=1}^m$ be a sample of $m$ i.i.d.\ examples
  from $D$. Suppose the loss $\ell \colon \R \times \R \to \R_+$ is
  $M$-bounded and $\mu$-Lipschitz with respect to its first
  argument. Fix $\delta > 0$. Then, the following hold:
\begin{enumerate}
\item For all $n \geq 1$ and $p \in [1,+\infty]$, with probability at least $1 - \delta$ the following holds simultaneously for all $A \in \cA_{n,p,1}$:
\begin{equation*}
\fL_D(A) \leq \hfL_S(A) + \sqrt{\frac{8 \mu^2 n(k n+2) \log(m+2)}{m}} + \frac{2 \mu (L_S + 2) }{m} + 3 M \sqrt{\frac{\log (2/\delta)}{2 m}} \enspace.
\end{equation*}
\item For all $r > 0$, with probability at least $1 - \delta$ the following holds simultaneously for all $f \in \cR_{1,r}$:
\begin{equation*}
\fL_D(f) \leq \hfL_S(f) + 
\frac{2 \mu r \sqrt{2 C_S \log (2m)}}{m}
+ 3 M \sqrt{\frac{\log (2/\delta)}{2 m}} \enspace.
\end{equation*}
\item For all $r > 0$, with probability at least $1 - \delta$ the following holds simultaneously for all $f \in \cH_{1,r}$:
\begin{align*}
\fL_D(f) \leq \hfL_S(f) &+ \left(\sqrt{2} + \frac{4}{\sqrt{\log 2}}\right) \frac{2 \mu r \sqrt{ W_S \log(2 m+1) }}{m} \\
&+ \left(\frac{2}{3} + \frac{8}{3 \log 2}\right) \frac{2 \mu r \log(2 m+1)}{m} + 3 M \sqrt{\frac{\log (2/\delta)}{2 m}} \enspace.
\end{align*}
\end{enumerate}
\end{theorem}

\section{Conclusion}
\label{sec:conclusion}

We presented the first algorithm-independent generalization bounds for
learning with wide classes of WFAs. We introduced three ways to
parametrize the complexity of WFAs and rational functions, each
described by a different natural quantity associated with the
automaton or function. We pointed out the merits of each description
in the analysis of the problem of learning with WFAs, and proved upper
bounds on the Rademacher complexity of several classes defined in terms
of these parameters.  An interesting property of these bounds is the
appearance of different combinatorial parameters that tie the sample
to the convergence rate: the length of the longest string $L_S$ for
$\cAnr$; the maximum number of collisions $C_S$ for $\cRpr$; and, the
minimum number of prefix or suffix collisions over all possible splits
$W_S$ for $\cHpr$.

Another important feature of our bounds for the classes $\cHpr$ is
that they depend on spectral properties of Hankel matrices, which are
commonly used in spectral learning algorithms for WFAs
\cite{hsu09,balle2012spectral}. We hope to exploit this connection in
the future to provide more refined analyses of these learning
algorithms.  Our results can also be used to improve some aspects of
existing spectral learning algorithms. For example, it might be
possible to use the analysis of Theorem~\ref{thm:Hpr} for deriving
strategies to help choose which prefixes and suffixes to consider in
algorithms working with finite sub-blocks of an infinite Hankel
matrix. This is a problem of practical relevance when working with
large amounts of data which require balancing trade-offs between
computation and accuracy \cite{mlj13spectral}.

It is possible to see that through a standard argument about the risk
of the empirical risk minimizer, our generalization bounds can be used
to establish that samples of size polynomial in the relevant
parameters are enough to learn in all the classes
considered. Nonetheless, the computational complexity of learning from
such a sample might be hard, since we know this is the case for DFAs
and PFAs \cite{PittWarmuth1993,KearnsValiant94,chalermsook2014pre}. In
the case of DFAs, several authors have analyzed special cases which
are tractable in polynomial time (e.g.\ \cite{clark2004partially} show
DFAs are learnable from positive data generated by ``easy''
distributions, and \cite{trakhtenbrot1973finite} showed that exact
learning can be done efficiently when the sample contains short
witnesses distinguishing every pair of states). For PFAs, spectral
methods show that polynomial learnability is possible if a new parameter
related to spectral properties of the Hankel matrix is added to the
complexity \cite{hsu09}.\ignore{ in the worst case can be exponential (e.g.\
for noisy parities \cite{kearns1994learnability}), but this approach
leads to efficient learning for a large class of distributions.} In the
case of general WFAs, there is no equivalent result identifying settings
in which the problem is tractable. In \cite{balle2012spectral}, we
proposed an efficient algorithm for learning WFAs that works in two
steps: a matrix completion procedure applied to Hankel matrices
followed by a spectral method to obtain a WFA from such Hankel
matrix. Although each of these two steps solves an optimization
problem without local minima, it is not clear from the analysis that
the solution of the combined procedure is close to the empirical risk
minimizer of any of the classes introduced in this paper. Nonetheless,
we expect that the tools developed in this paper will prove useful in
analyzing variants of this algorithm and will also help design
new algorithms for efficiently learning interesting classes of WFA.

%\subsection*{Acknowledgments}
%
%This work was partly funded by  BLA BLA

%\section*{References}

%\bibliographystyle{elsarticle-num}
%\bibliographystyle{model1-num-names}
\bibliographystyle{plain}
\bibliography{lwfa}

\begin{thebibliography}{10}

\bibitem{abe1992computational}
Naoki Abe and Manfred~K Warmuth.
\newblock On the computational complexity of approximating distributions by
  probabilistic automata.
\newblock {\em Machine Learning}, 1992.

\bibitem{AllauzenMohriRiley2004}
Cyril Allauzen, Mehryar Mohri, and Michael Riley.
\newblock Statistical modeling for unit selection in speech synthesis.
\newblock In {\em Proceedings of ACL}, 2004.

\bibitem{AllauzenMohriTalwalkar2008}
Cyril Allauzen, Mehryar Mohri, and Ameet Talwalkar.
\newblock Sequence kernels for predicting protein essentiality.
\newblock In {\em Proceedings of ICML}, 2008.

\bibitem{AminofKupfermanLampert2011}
Benjamin Aminof, Orna Kupferman, and Robby Lampert.
\newblock Formal analysis of online algorithms.
\newblock In {\em Proceedings of ATVA}, 2011.

\bibitem{baier2009model}
C.~Baier, M.~Gr{\"o}{\ss}er, and F.~Ciesinski.
\newblock Model checking linear-time properties of probabilistic systems.
\newblock In {\em Handbook of Weighted automata}. Springer, 2009.

\bibitem{denis}
R.~Bailly, F.~Denis, and L.~Ralaivola.
\newblock Grammatical inference as a principal component analysis problem.
\newblock In {\em ICML}, 2009.

\bibitem{decideabsconv}
Rapha{\"{e}}l Bailly and Fran{\c{c}}ois Denis.
\newblock Absolute convergence of rational series is semi-decidable.
\newblock {\em Inf. Comput.}, 2011.

\bibitem{mlj13spectral}
B.~Balle, X.~Carreras, F.M. Luque, and A.~Quattoni.
\newblock Spectral learning of weighted automata: A forward-backward
  perspective.
\newblock {\em Machine Learning}, 2014.

\bibitem{icml2014balle}
B.~Balle, W.L. Hamilton, and J.~Pineau.
\newblock Methods of moments for learning stochastic languages: Unified
  presentation and empirical comparison.
\newblock In {\em ICML}, 2014.

\bibitem{balle2012spectral}
Borja Balle and Mehryar Mohri.
\newblock Spectral learning of general weighted automata via constrained matrix
  completion.
\newblock In {\em NIPS}, 2012.

\bibitem{cai2015}
Borja Balle and Mehryar Mohri.
\newblock Learning weighted automata.
\newblock In {\em CAI}, 2015.

\bibitem{bpp15}
Borja Balle, Prakash Panangaden, and Doina Precup.
\newblock A canonical form for weighted automata and applications to
  approximate minimization.
\newblock In {\em Logic in Computer Science (LICS)}, 2015.

\bibitem{bartlett2001rademacher}
Peter~L Bartlett and Shahar Mendelson.
\newblock Rademacher and gaussian complexities: Risk bounds and structural
  results.
\newblock In {\em COLT}, 2001.

\bibitem{BerstelReutenauer1988}
Jean Berstel and Christophe Reutenauer.
\newblock {\em Rational Series and Their Languages}.
\newblock Springer, 1988.

\bibitem{berstel2011noncommutative}
Jean Berstel and Christophe Reutenauer.
\newblock {\em Noncommutative rational series with applications}.
\newblock Cambridge University Press, 2011.

\bibitem{Boots:2009}
B.~Boots, S.~Siddiqi, and G.~Gordon.
\newblock Closing the learning-planning loop with predictive state
  representations.
\newblock In {\em RSS}, 2009.

\bibitem{Breuel2008}
Thomas~M. Breuel.
\newblock The {OCR}opus open source {OCR} system.
\newblock In {\em Proceedings of IS\&T/SPIE}, 2008.

\bibitem{CarlylePaz1971}
Jack~W. Carlyle and Azaria Paz.
\newblock Realizations by stochastic finite automata.
\newblock {\em J. Comput. Syst. Sci.}, 5(1), 1971.

\bibitem{chalermsook2014pre}
P.~Chalermsook, B.~Laekhanukit, and D.~Nanongkai.
\newblock Pre-reduction graph products: Hardnesses of properly learning dfas
  and approximating edp on dags.
\newblock In {\em Proceedings of FOCS}, 2014.

\bibitem{clark2004partially}
Alexander Clark and Franck Thollard.
\newblock Partially distribution-free learning of regular languages from
  positive samples.
\newblock In {\em Proceedings of the 20th international conference on
  Computational Linguistics}, page~85. Association for Computational
  Linguistics, 2004.

\bibitem{CortesHaffnerMohri2004}
Corinna Cortes, Patrick Haffner, and Mehryar Mohri.
\newblock Rational kernels: Theory and algorithms.
\newblock {\em Journal of Machine Learning Research}, 5, 2004.

\bibitem{cortes2007lp}
Corinna Cortes, Mehryar Mohri, and Ashish Rastogi.
\newblock Lp distance and equivalence of probabilistic automata.
\newblock {\em International Journal of Foundations of Computer Science}, 2007.

\bibitem{DeGispert2010}
A.~de~Gispert, G.~Iglesias, G.~Blackwood, E.R. Banga, and W.~Byrne.
\newblock Hierarchical phrase-based translation with weighted finite-state
  transducers and shallow-n grammars.
\newblock {\em Computational Linguistics}, 2010.

\bibitem{devroye2001combinatorial}
Luc Devroye and G{\'a}bor Lugosi.
\newblock {\em Combinatorial methods in density estimation}.
\newblock Springer, 2001.

\bibitem{DrosteKuich2009}
Manfred Droste, Werner Kuich, and Heiko Vogler, editors.
\newblock {\em Handbook of weighted automata}.
\newblock {EATCS} Monographs on Theoretical Computer Science. Springer, 2009.

\bibitem{dudley1999uniform}
Richard~M Dudley.
\newblock {\em Uniform central limit theorems}, volume~23.
\newblock Cambridge Univ Press, 1999.

\bibitem{DurbinEddyKroghMitchison1998}
Richard Durbin, Sean~R. Eddy, Anders Krogh, and Graeme~J. Mitchison.
\newblock {\em Biological Sequence Analysis: Probabilistic Models of Proteins
  and Nucleic Acids}.
\newblock Cambridge University Press, 1998.

\bibitem{Eilenberg1974}
Samuel Eilenberg.
\newblock {\em Automata, Languages and Machines}, volume~A.
\newblock Academic Press, 1974.

\bibitem{Fliess1974}
M.~Fliess.
\newblock Matrices de {H}ankel.
\newblock {\em Journal de Math\'ematiques Pures et Appliqu\'ees}, 53, 1974.

\bibitem{Hamilton:2013}
W.~L. Hamilton, M.~M. Fard, and J.~Pineau.
\newblock Modelling sparse dynamical systems with compressed predictive state
  representations.
\newblock In {\em ICML}, 2013.

\bibitem{hsu09}
D.~Hsu, S.~M. Kakade, and T.~Zhang.
\newblock A spectral algorithm for learning hidden {M}arkov models.
\newblock In {\em COLT}, 2009.

\bibitem{CulikIIKari1993}
Karel~Culik II and Jarkko Kari.
\newblock Image compression using weighted finite automata.
\newblock {\em Computers {\&} Graphics}, 17(3), 1993.

\bibitem{Ishigami1997123}
Yoshiyasu Ishigami and Sei'ichi Tani.
\newblock Vc-dimensions of finite automata and commutative finite automata with
  k letters and n states.
\newblock {\em Discrete Applied Mathematics}, 1997.

\bibitem{KaplanKay1994}
Ronald~M. Kaplan and Martin Kay.
\newblock Regular models of phonological rule systems.
\newblock {\em Computational Linguistics}, 20(3), 1994.

\bibitem{Karttunen1995}
Lauri Karttunen.
\newblock The replace operator.
\newblock In {\em Proceedings of ACL}, 1995.

\bibitem{KearnsValiant94}
Michael~J. Kearns and Leslie~G. Valiant.
\newblock Cryptographic limitations on learning boolean formulae and finite
  automata.
\newblock {\em Journal of ACM}, 41(1), 1994.

\bibitem{koltchinskii2000rademacher}
Vladimir Koltchinskii and Dmitry Panchenko.
\newblock Rademacher processes and bounding the risk of function learning.
\newblock In {\em High Dimensional Probability II}, pages 443--459.
  Birkh\"{a}user, 2000.

\bibitem{KuichSalomaa1986}
Werner Kuich and Arto Salomaa.
\newblock {\em Semirings, Automata, Languages}.
\newblock Number~5 in {EATCS} Monographs on Theoretical Computer Science.
  Springer-Verlag, Berlin-New York, 1986.

\bibitem{kulesza2015low}
A.~Kulesza, N.~Jiang, and S.~Singh.
\newblock Low-rank spectral learning with weighted loss functions.
\newblock In {\em AISTATS}, 2015.

\bibitem{kulesza2014low}
Alex Kulesza, N~Raj Rao, and Satinder Singh.
\newblock {Low-Rank Spectral Learning}.
\newblock In {\em AISTATS}, 2014.

\bibitem{ledtal}
Michel Ledoux and Michel Talagrand.
\newblock {\em Probability in Banach spaces}.
\newblock Springer-Verlag, 1991.

\bibitem{Massart2000}
Pascal Massart.
\newblock Some applications of concentration inequalities to statistics.
\newblock {\em Annales de la Facult\'e des Sciences de Toulouse}, 2000.

\bibitem{vonneumann}
L.~Mirsky.
\newblock A trace inequality of {John von Neumann}.
\newblock {\em Monatshefte für Mathematik}, 1975.

\bibitem{Mohri1997}
Mehryar Mohri.
\newblock Finite-state transducers in language and speech processing.
\newblock {\em Computational Linguistics}, 23(2), 1997.

\bibitem{Mohri2009}
Mehryar Mohri.
\newblock Weighted automata algorithms.
\newblock In {\em Handbook of Weighted Automata}, Monographs in Theoretical
  Computer Science, pages 213--254. Springer, 2009.

\bibitem{MohriPereiraRiley1996}
Mehryar Mohri, Fernando Pereira, and Michael Riley.
\newblock Weighted automata in text and speech processing.
\newblock In {\em Proceedings of ECAI-96 Workshop on Extended finite state
  models of language}, 1996.

\bibitem{MohriPereira1998}
Mehryar Mohri and Fernando C.~N. Pereira.
\newblock Dynamic compilation of weighted context-free grammars.
\newblock In {\em Proceedings of COLING-ACL}, 1998.

\bibitem{MohriPereiraRiley2008}
Mehryar Mohri, Fernando C.~N. Pereira, and Michael Riley.
\newblock Speech recognition with weighted finite-state transducers.
\newblock In {\em {Handbook on Speech Processing and Speech Comm.}} Springer,
  2008.

\bibitem{mohri2012foundations}
Mehryar Mohri, Afshin Rostamizadeh, and Ameet Talwalkar.
\newblock {\em Foundations of machine learning}.
\newblock MIT press, 2012.

\bibitem{MohriSproat1996}
Mehryar Mohri and Richard Sproat.
\newblock An efficient compiler for weighted rewrite rules.
\newblock In {\em Proceedings of ACL}, 1996.

\bibitem{PereiraRiley1997}
Fernando Pereira and Michael Riley.
\newblock Speech recognition by composition of weighted finite automata.
\newblock In {\em Finite-State Language Processing}. {MIT} Press, 1997.

\bibitem{PittWarmuth1993}
Leonard Pitt and Manfred~K. Warmuth.
\newblock The minimum consistent {DFA} problem cannot be approximated within
  any polynomial.
\newblock {\em Journal of the ACM}, 40(1), 1993.

\bibitem{SalomaaSoittola1978}
Arto Salomaa and Matti Soittola.
\newblock {\em Automata-Theoretic Aspects of Formal Power Series}.
\newblock Springer-Verlag: New York, 1978.

\bibitem{Sproat1995}
Richard Sproat.
\newblock A finite-state architecture for tokenization and grapheme-to-phoneme
  conversion in multilingual text analysis.
\newblock In {\em Proceedings of the ACL SIGDAT Workshop}. ACL, 1995.

\bibitem{trakhtenbrot1973finite}
B~Trakhtenbrot and Y~Barzdin.
\newblock {\em Finite Automata: Behavior and Synthesis}.
\newblock North-Holland, 1973.

\bibitem{tropp2015introduction}
Joel~A. Tropp.
\newblock An introduction to matrix concentration inequalities.
\newblock {\em Foundations and Trends{\textregistered} in Machine Learning},
  8(1-2):1--230, 2015.

\bibitem{vershynin-gfa}
Roman Vershynin.
\newblock {Lectures in Geometrical Functional Analysis}.
\newblock {\em Preprint}, 2009.

\end{thebibliography}

\end{document}